\documentclass[lettersize,journal]{IEEEtran}
\usepackage[english]{babel}
\usepackage{amsmath,amsfonts}
\usepackage{amssymb}
\usepackage{algorithmic}
\usepackage{algorithm}
\usepackage{array}
\usepackage{textcomp}
\usepackage{stfloats}
\usepackage{url}
\usepackage{verbatim}
\usepackage{graphicx}
\usepackage{cite}
\usepackage[table,xcdraw]{xcolor}
\usepackage{multirow}
\usepackage{subcaption}
\usepackage[colorlinks, citecolor=cyan]{hyperref}

\begin{document}

\title{Truck Parking Usage Prediction with Decomposed Graph Neural Networks}

\author{Rei Tamaru, Yang Cheng, Steven Parker, Ernie Perry, Bin Ran, and Soyoung Ahn
\thanks{The Truck Parking Information System (TPIMS) was developed through sponsorship and collaboration with Mid America Freight Coalition (MAFC) and participating states of Mid America Association of State Transportation Officials (MAASTO). The ideas and views expressed in this paper are strictly those of the Traffic Operations and Safety (TOPS) Laboratory at the University of Wisconsin-Madison.}
\thanks{R. Tamaru, Y. Cheng, S. Parker, E. Perry, B. Ran, and S. Ahn are with the Department of Civil and Environmental Engineering, University of Wisconsin-Madison, Madison, Wisconsin, United States, 53705 (email: tamaru@wisc.edu; cheng8@wisc.edu; sparker@engr.wisc.edu; ebperry@wisc.edu; bran@wisc.edu; sue.ahn@wisc.edu;)}}


\maketitle

\begin{abstract}
Truck parking on freight corridors faces the major challenge of insufficient parking spaces. This is exacerbated by the Hour-of-Service (HOS) regulations, which often result in unauthorized parking practices, causing safety concerns. It has been shown that providing accurate parking usage prediction can be a cost-effective solution to reduce unsafe parking practices. In light of this, existing studies have developed various methods to predict the usage of a truck parking site and have demonstrated satisfactory accuracy. However, these studies focused on a single parking site, and few approaches have been proposed to predict the usage of multiple truck parking sites considering spatio-temporal dependencies, due to the lack of data. This paper aims to fill this gap and presents the Regional Temporal Graph Convolutional Network (RegT-GCN) to predict parking usage across the entire state to provide more comprehensive truck parking information. The framework leverages the topological structures of truck parking site locations and historical parking data to predict the occupancy rate considering spatio-temporal dependencies across a state. To achieve this, we introduce a Regional Decomposition approach, which effectively captures the geographical characteristics of the truck parking locations and their spatial correlations. Evaluation results demonstrate that the proposed model outperforms other baseline models, showing the effectiveness of our regional decomposition. The code is available at \url{https://github.com/raynbowy23/RegT-GCN}.
\end{abstract}


\section{Introduction}
The Federal Motor Carrier Safety Administration (FMCSA) has implemented regulations to reduce fatigue-related safety hazards for truck drivers by restricting the consecutive and total daily driving hours permitted for truck operators. Despite these efforts, truck drivers still spend considerable time searching for safe parking locations beyond their limited legal working hours \cite{boris2018}. This phenomenon highlights the persistent issue of inadequate truck parking facilities, contributing to unauthorized parking behaviors and breaches of parking regulations \cite{fhwa2015}.

The literature identifies increasing truck drivers' accessibility to real-time parking usage information \cite{smith2005} and drivers' perceptual interpretations \cite{gates2012, maze2010} as major factors influencing such parking challenges. Along with the survey studies in the region \cite{maasto2018}, members of the Mid-America Association of State Transportation Officials (MAASTO) from eight states, Indiana, Iowa, Kansas, Kentucky, Michigan, Minnesota, Ohio, and Wisconsin, have collaborated to build a real-time multistate Truck Parking Information Management System (TPIMS) to provide more practical truck parking information to truck drivers \cite{maasto2016, moore2018}. TPIMS, fully operational since January 4, 2019, provides real-time parking usage information to truck drivers through dynamic message signs, smartphone applications, traveler information websites (e.g. 511 traveler information) and other forms.

In addition to real-time parking site availability information, future usage prediction can better support truck drivers in their decision-making process and route planning. Current prediction models exhibit commendable accuracy in predicting future usage for the subsequent hour \cite{sadek2020, hao2022}. However, it is crucial to recognize that the domain of interest encompasses multiple truck parking sites situated along the highway corridor. Boris and Brewster \cite{boris2018} reported that the most critical aspects of the target parking sites are the proximity to the route/destination followed by the availability of amenities. Therefore, the primary objective of our prediction model is to capitalize on the wealth of data available from various truck parking sites and leverage their topological structures. 


\begin{figure*}[!ht]
    \centering
    \includegraphics[width=0.9\textwidth]{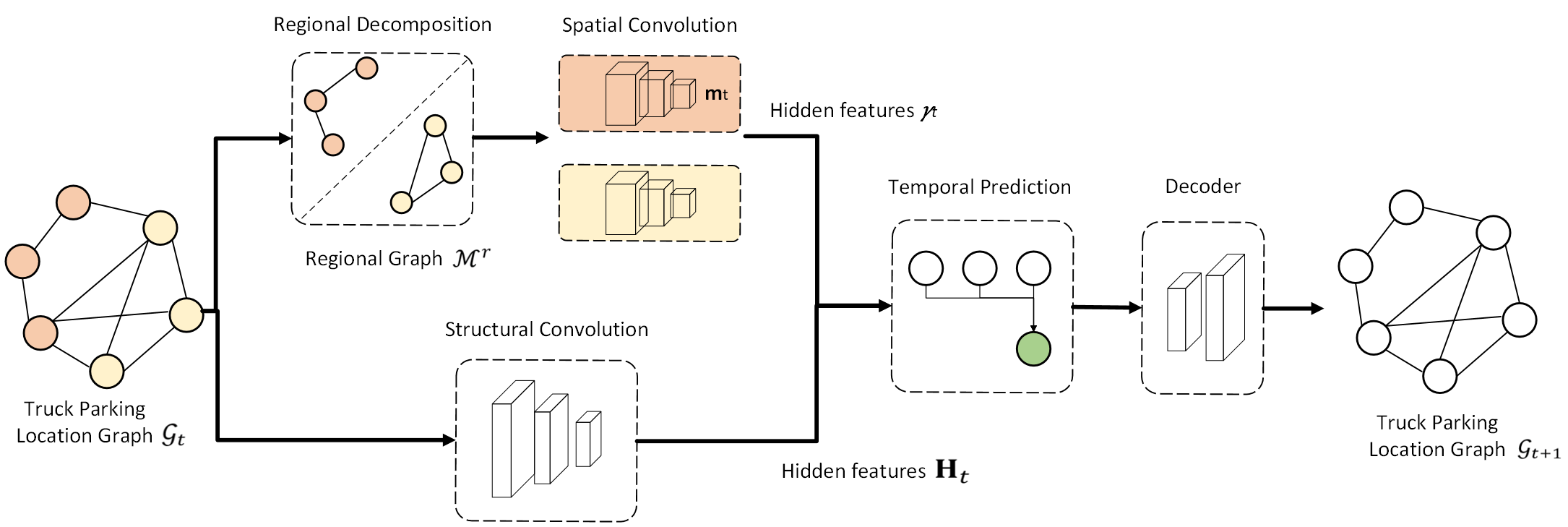}
    \caption{The overview of Regional Temporal Graph Neural Network.}\label{fig:overview}
\end{figure*}

To this end, we aim to develop a data-driven, predictive model for future parking usage, grounded in spatio-temporal dependencies of parking sites on highway corridors and their historical usage patterns. In this research, we propose a Regional Temporal Graph Convolutional Network (RegT-GCN), comprising a Graph Convolutional Network (GCN) with temporal capability and a novel decomposition module, Regional Decomposition, to capture regional and spatial relationships (Figure \ref{fig:overview}). In this study, a region is defined as a state to incorporate their unique characteristics (e.g., state laws). The significant historical parking data allows the model to capture temporal dependencies of multi-state-wide truck parking usage to generate more accurate prediction results. We conduct several experiments to evaluate the model against various models and raised the importance of Regional Decomposition technique for Graph Convolutional Networks.

Our contributions are summarized as follows.

\begin{itemize}
\item We propose a spatio-temporal model to predict the usage of multiple truck parking sites. 
\item Our novel Regional Decomposition method leverages the regional relationships to create subgraphs for each region (i.e., state) and effectively captures the spatio-temporal dependencies.
\item We construct the comprehensive truck parking dataset, which is aggregated across multi-states to train their spatio-temporal dependencies and evaluate the prediction accuracy of occupancy rates.
\item We conduct extensive quantitative evaluations including temporal predictions, spatial analysis, and sensitivity analysis, which demonstrate that our RegT-GCN significantly outperforms baseline models.
\end{itemize}

\section{Literature Review}

\subsection{Spatio-Temporal Prediction in Transportation}
Spatio-temporal dependencies in the transportation field are notably effective for traffic flow prediction as they take advantage of highway network topology. Those approaches adopted graph convolutions \cite{zhao2020, liang2023, bao2023}, attention mechanism \cite{jiang2023, chen2024}, and transformer \cite{qinyao2024}. Specifically, GCNs \cite{kipf2016} are highly effective in capturing spatial embeddings used with temporal embedding modules \cite{zhao2020, bao2023} while the attention mechanism improves its performance in long-range temporal dependencies working in conjunction with spatial embedding models \cite{jiang2023, qinyao2024}. Recent works such as STID \cite{shao2022} and STAEformer \cite{liu2023} extended spatial-temporal dependencies with more sophisticated embedding networks. All these methods, however, lack a deep investigation of spatial embeddings that align with urban and truck parking usage prediction.

\subsection{Parking Usage Prediction}
Parking prediction methods can be classified into two main categories: urban and truck parking prediction. Urban parking prediction typically offers a shorter horizon, ranging from 5 to 15 minutes, while truck parking prediction extends more than an hour ahead \cite{hao2022}. Despite these differences in time frames, both tasks share underlying concepts and are designed to address a variety of parking scenarios. Models that operate parking prediction combine historical data and real-time updates \cite{bayraktar2015, haque2017}. This is extended in the pattern recognition approach \cite{sadek2020, hamidreza2019}. Vital et al. \cite{vital2021} further enhanced prediction accuracy by capturing spatio-temporal dependencies. Recent studies \cite{zhao2020, zheng2020, low2020} have adopted machine learning techniques to integrate additional features, such as weather, daily patterns, and parking location. Although feature embeddings are effective for parking prediction \cite{hao2022}, these studies limit the prediction capability within a few sites.

\subsection{Spatio-Temporal Representation in Parking Prediction}
While most prior research on predicting truck parking utilization focuses primarily on historical parking usage data and static variables, it overlooked the inclusion of spatial dependencies as the key determinants of truck driver's decision-making process \cite{boris2018}. This gap highlights the need for integrating topological structures into predictive models to account for spatial variations in influential factors.

This need for spatial integration aligns with the road network's inherent structure, which naturally functions as a non-Euclidean graph \cite{jiang2022}. Yang et al. \cite{yang2019} first leveraged GCN on urban parking to extract the spatial dependencies of traffic flow in large-scale networks and incorporated Recurrent Neural Networks (RNN) and Long Short Term Memory (LSTM) to capture temporal dependencies with additional related features. Zhang et al. \cite{zhang2020} addressed the scarcity of real-time parking availability information with a semi-supervised approach and spatio-temporal dependencies among parking lots. Despite their success in urban parking with spatio-temporal dependencies, they have yet to conduct modeling and experiments on truck parking usage prediction due to the lack of sufficient data.

Therefore, our method benefits from the model with spatio-temporal embeddings while adapting truck parking unique features such as amenities, capacity limits \cite{sadek2020}, and diurnal patterns \cite{xiao2023}. In contrast to the introduction of the spatio-temporal prediction model in an urban parking scenario \cite{gong2023}, we consider the unique regional dependencies to capture a large truck parking network.



\section{Methodological Approach}

\subsection{Problem Formulation} \label{problem_form}
We describe the problem formulation for the task of predicting future truck parking usage, as a novel model to represent spatio-temporal dependencies on the parking. Starting from the general notation, a truck parking graph is denoted as $\mathcal{G} = (\mathcal{V}, \mathcal{E})$, where individual parking sites are represented as nodes $\mathbf{v} \in \mathcal{V}$, and the connections between these sites are represented as edges $\mathbf{e} \in \mathcal{E}$. Within this framework, we denote node-embedding as $\mathbf{\eta}$, representing each site's characteristics of features numerically. 

Our primary challenge is to predict the occupancy rates of truck parking sites in the next $T$ future time intervals. To predict accurate future occupancy rates $o$ at the site $i$, we incorporate information from the $K$ preceding consecutive time intervals. This approach enables us to capture temporal patterns in the use of truck parking.

Our prediction task is formulated by the equation:
\begin{equation}
    \eta_{i_{t-K}}, \ldots, \eta_{i_t} \rightarrow \hat \eta_{i_{t+1}}, \ldots, \hat \eta_{i_{t+T}}
\end{equation}

In this equation, the left-hand side represents the historical node embeddings for a specific parking site up to time $t$. These embeddings encode the site's historical characteristics. Our model then processes this information to generate predictions for the occupancy rates in the subsequent time intervals, shown on the right-hand side. These predicted embeddings represent future occupancy rates.

Then, we generalize the prediction target across all parking sites within our dataset, solving the prediction task:
\begin{equation}
    [\mathcal{V}_{t-K}, \ldots, \mathcal{V}_t; \mathcal{E}; \mathcal{G}] \rightarrow [\mathcal{\hat V}_{t+1}, \ldots, \mathcal{\hat V}_{t+T}]
\end{equation}

In order to capture the spatio-temporal dependencies, this research first engages in understanding regional relationships across truck parking sites. Based on our preliminary analysis \cite{rei2023} and insights from previous studies on the efficacy of edge reduction techniques \cite{rong2019, hamilton2017}, we assume that the regional relationship within the truck parking network can be leveraged to reduce graph complexity. We hypothesize that a regionally decomposed graph structure, as opposed to one connected graph, offers advantages in terms of both predictive performance and computational efficiency.

In this paper, we define small partitions in the truck parking sites as an independent \textit{region} and explore the spatial dependencies among truck parking sites. Site-specific parking behavior deliberates region selection to be grounded in practicality and the unique characteristics exhibited by states \cite{nchrp2003}. For example, freight plans vary by state, so they recommend different values and strategies for developing, funding, and maintaining shared facilities. It is also beneficial to consider the state as a region from a data aggregation perspective. Hence, our regional selection prioritizes interpretability and practical outcomes.

Formally, we define regional graphs as subgraphs of truck parking graphs, denoted as $\mathcal{M} = (\mathcal{V'}, \mathcal{E'})$ segmented according to regional areas, where $\mathcal{M} \neq \mathcal{G}, \mathcal{M} \subseteq \mathcal{G}$. Each regional graph $\mathcal{M}^r \subseteq \mathcal{M}$ consists of a subset of vertices and edges $\mathcal{M}^r = \{(\mathcal{V}^r, \mathcal{E}^r) | \forall r \in \text{region}\}$. The adjacency matrix of the overall graph $\mathcal{G}$ is represented as $\mathbf{A} \in \mathbb{R}^{N \times N}$ and can be defined as follows, where the vertices of $\mathcal{G}$ are labeled $\mathbf{v_1}, \ldots, \mathbf{v_n}$ and the entries are binary (${0, 1}$).

\begin{equation}
    \mathbf{e}_{ij} = \mathbf{v}_i \mathbf{v}_j \in \mathcal{E} \iff \mathbf{A}_{ij} = 1
\end{equation}

Similarly, the adjacency matrix of the regional subgraph $\mathcal{M}^r \subseteq \mathcal{M}$ denoted as $\mathbf{A}^r \subseteq \mathbf{A}$, is selectively defined for each state in MAASTO as follows.

\begin{equation}
    \mathbf{e}^r_{ij} = \mathbf{v}^r_i \mathbf{v}^r_j \in \mathcal{E}^r \iff \mathbf{A}^r_{ij} = 1
\end{equation}

This segregation of the graph into regional subgraphs reduces the complexity of the network according to the degrees of the nodes, denoted as $d^i_r$ for site $i$:

\begin{equation}
    \mathbf{d}^r_i = \sum_{j \in \mathcal{V}^r} \mathbf{A}^r_{ij} \subseteq \sum_{j\in \mathcal{V}} \mathbf{A}_{ij}
\end{equation}

Therefore, our framework incorporates the invention of graph decomposition strategies with the construction of subgraphs by deliberately selecting nodes from regional sets within states in the MAASTO. Our prediction task is now defined as follows.

\begin{equation}
\begin{split}
    [\mathcal{V}_{t-K}, \ldots, \mathcal{V}_t; \mathcal{E}; \mathcal{G}] &= \sigma \bigcup^r[\mathcal{V}^r_{t-K+1}, \ldots, \mathcal{V}^r_t; \mathcal{E}^r; \mathcal{M}^r] \\
                                 &\rightarrow \bigcup^r[\mathcal{\hat V}^r_{t+1}, \ldots, \mathcal{\hat V}^r_{t+T}]
\end{split}
\end{equation}
where $\sigma$ represents a non-linear activation function such as ReLU and tanh.




\subsection{Regional Temporal Graph Convolutional Network}

In this paper, we built the graph-based spatio-temporal model, which incorporates the geographical characteristics in nodes integrating spatial and temporal aspects with regional graph input. Our model comprises two main components: a GCN, inspired by the Attention Temporal Graph Convolutional Network (A3TGCN) \cite{bai2021}, as a spatial module and a recurrent unit as a temporal module. A structual GCN handles the node embeddings of the truck parking location graph, and a spatial GCN compresses the spatial dependencies between sites taken the distances between each pair of truck parking sites. Given the sequence of input timestamps $t \in \{1,2, \ldots, K\}$, the temporal model predicts site occupancy rates up to $T$ future time steps.


For each time step $t$, in the node embedding at a graph level, we denote $\mathbf{\hat{A}}$ as the adjacency matrix with self-loop, where $\mathbf{\hat{A}} = \mathbf{A} + \mathbf{I}$ with $\mathbf{I}$ being the identity matrix. Additionally, $\mathbf{\hat D}$ represents a diagonal degree matrix, its elements are defined as $\mathbf{\hat{D}}_{ii} = \sum_{j}\mathbf{\hat{A}}_{ij}$ and the weight matrix is denoted as $\mathbf{W}$. It demands $\mathbf{H}_{t+1}$ of the hidden vectors. Following the equation proposed by Kipf and Welling \cite{kipf2016}, this progression can be expressed as:
\begin{equation}
    \mathbf{H}_{t+1} = \sigma (\mathbf{\hat D}^{-\frac{1}{2}} \mathbf{\hat A} \mathbf{\hat D}^{-\frac{1}{2}}\mathbf{H}_t\mathbf{W}_t)
\end{equation}




\begin{algorithm}[t] 
\caption{Algorithm overview of whole procedure} \label{alg:whole}
\begin{algorithmic}
\REQUIRE $\mathcal{G} = (\mathcal{V}, \mathcal{E}), T \geq 0, K \geq 0$
\FOR{$t=1$ to $T$}
   \STATE $\mathcal{G}_t = (\mathcal{V}_t, \mathcal{E}_t)$;
   
   \FOR{$k = 1$ to $K$}
       \STATE $\mathbf{H}_{t-k} \gets \text{Structural GCN} (\mathcal{G}_{t-k})$
        \STATE $\mathcal{M}_{t-k} \gets \text{Regional Decomposition} (\mathcal{G}_{t-k})$
        \FOR{$r$ \text{in states in MAASTO}}
            \STATE $\mathbf{m}^r_{t-k} \gets \text{Spatial GCN} (\mathcal{M}^r_{t-k})$
        \ENDFOR
        \STATE $\mathbf{\gamma}_{t-k} \gets \bigoplus_r \mathbf{m}^r_{t-k}$
    \ENDFOR

    \STATE $\mathbf{H}_{t+1} \gets \sum^K_{k=1} \alpha_t \text{Temporal}(\mathbf{H}_{t-k}, \mathbf{\gamma}_{t-k})$
    \STATE $\mathcal{V}_{t+1} \gets \text{Decode} (\mathbf{H}_{t+1})$
\ENDFOR
   
\end{algorithmic} 
\end{algorithm}

To leverage the potential benefits of regional relationships, we introduce the Regional Decomposition technique. This involves the creation of subgraphs $\mathcal{M}$ while decomposing the original graph of the truck parking site locations $\mathcal{G}$. During each training step, the Regional Decomposition process distributes each node attribute $\mathbf{\eta}$ by regional relationships. This process leads to deriving independent vectors $\{\mathbf{m}_1, \ldots, \mathbf{m}_N\} \in \mathbb{R}^{N \times E}$ with the number of features $E$ and the total number of truck parking site $N$.

The input graphs are sequentially aligned at every 10-minute interval and passed to spatial modules to extract the node features. Each feature represents structural information from the regional graph and information from the nodes based on features within the regional context. Our whole algorithm is shown in Algorithm \ref{alg:whole}. Each GCN is applied to the regional graphs to extract spatial features as hidden vectors $\mathbf{m}$. Subsequently, these vectors are concatenated, and the linear layer computes the hidden vector $\mathbf{\gamma}$, which can be used in the temporal layers. These operations are performed iteratively for each time step.

Within the spatial module, two distinct categories of spatial models can be identified: the structural model and the model for node embeddings. The structural model serves as a means to comprehend the composition of graphs, while the spatial model is employed to extract node attributes within temporal layers.

Formally, we arbitrarily collect the region of the nodes in $\mathcal{G}$ and enforce them to connect together as a graph $\mathcal{M}^r$. On the subgraph networks, we can consider $\mathbf{\gamma}$ as a concatenated feature derived by the node embeddings from spatial GCN denoted as
\begin{equation}
\mathbf{\gamma} = \bigcup_{r} (\mathbf{m}^r, \mathbf{e}^{r}) 
\end{equation}

\subsection{Temporal Module}

\begin{figure}[!ht]
    \centering
    \includegraphics[width=0.6\linewidth]{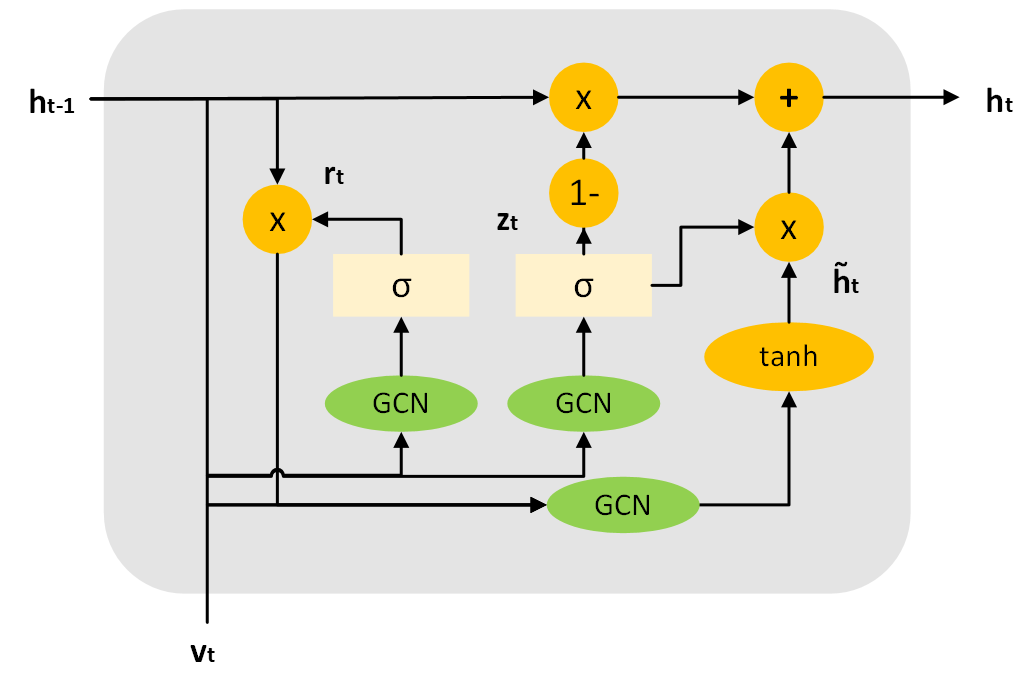}
    \caption{The architecture of GCN based GRU.}\label{fig:gru_gcn}
\end{figure}

The temporal model is applied to acquire temporal dependencies with hidden vectors from the previous time step. To obtain the temporal dependencies of composed features, we constructed the GRU architecture using GCN, inspired by the baseline model \cite{bai2021}. This is the extension technique of Convolutional LSTM \cite{shi2015} to handle spatial dependencies into a temporal module. Figure \ref{fig:gru_gcn} illustrates the architecture of GRU, which can accommodate the graph structure and the composed features. Within this framework, the activation function $\sigma$ represents the sigmoid function, and the parameters $\mathbf{W}$ correspond to the weights. The function $f(\mathbf{v}_t, \mathbf{A}_t)$ denotes the computation of graph convolutions in the time step $t$, which is subsequently concatenated with the input hidden vectors. Accordingly, the structural convolution function can be formally defined as follows:

\begin{equation}
    f(\mathbf{v}, \mathbf{A}) = \sigma(\mathbf{W} \gamma_i + \mathbf{W} \sum^{\mathbf{d}_i}_l \gamma_i^{(l)})
\end{equation}

Incorporating the structural formulas and the temporal dependency function, our RegT-GCN formulation at the node level can be expressed as follows with the initial condition of $\mathbf{h}_{0} = \gamma_{0}$.
\begin{equation}
\begin{split}
    \mathbf{z}_t &= \sigma (\mathbf{W}_z f(\mathbf{v}_t, \mathbf{A}_t)\oplus \mathbf{h}_{t-1}) \\
    \mathbf{r}_t &= \sigma (\mathbf{W}_r f(\mathbf{v}_t, \mathbf{A}_t)\oplus \mathbf{h}_{t-1}) \\
    \mathbf{\tilde{h}}_t &= tanh (\mathbf{W} f(\mathbf{v}_t, \mathbf{A}_t)\oplus (\mathbf{h}_{t-1} \odot \mathbf{r}_t)) \\
    \mathbf{h}_t &= (1-\mathbf{z}_t) \odot \mathbf{h}_{t-1} + \mathbf{z}_t \odot \tilde{\mathbf{h}_t}
\end{split}
\end{equation}
where $\oplus$ denotes the concatenation operation applied to the hidden vectors. This formulation enables the modeling of temporal dependencies and information propagation in the graph-based neural network.

\subsection{Model Training}
After iterating the input time steps in the temporal module, they are all passed to the decoder, consisting of two linear layers and the ReLU activation function. 

We also modified the attention vector $\alpha_t$ as an aggregation function in simplified form to improve computational efficiency and in accordance with our experimental decision. 

\begin{equation}
    \alpha_t = \frac{\exp(\alpha_t)}{\sum_{s \in \mathbb{N}^{K} \setminus \{t\}}\exp(\alpha_s)}
\end{equation}

Finally, the calculated hidden vectors are passed through the decoder to yield the predicted outputs for the nodes at the time step $t+1$, denoted $\mathbf{V}_{t+1}$, and the hidden vector matrix is expressed as $\mathbf{H}_{t}$.

\begin{equation}
\begin{split}
    \mathbf{V}_{t+1} &= \text{ReLU}(\mathbf{H}_{t} \mathbf{W}^{(0)} + \mathbf{b}^{(0)})\mathbf{W}^{(1)} + \mathbf{b}^{(1)} \\
        &= \text{ReLU}(\text{ReLU} (\alpha_t f(\mathbf{V}_t, \mathbf{A}_t) \oplus \mathbf{H}_{t-1}) \mathbf{W}^{(0)} \\
        &+ \mathbf{b}^{(0)})\mathbf{W}^{(1)} + \mathbf{b}^{(1)} 
\end{split}
\end{equation}

During the training process, the model is trained to minimize the discrepancy between the actual truck parking occupancy rates and predicted values across time steps. The loss of the model is then calculated using the target attribute, the occupancy rates matrix $O_t$, using the mean squared error criterion.

\begin{equation}
    \mathcal{L} = \frac{1}{N} \sum_{n=1}^{N} (\mathbf{O}_{n_t} - \mathbf{\hat{O}}_{n_t})^2
\end{equation}
where $\mathbf{\hat{O}}$ is the matrices of predicted occupancy rates.

\section{Experiments} \label{sec:experiments}

This section explains data, data handling, experiment settings, and the results. There are two perspectives of results, including truck parking usage prediction in different time horizons and the efficiency of graph connectivity regarding the model performance of Regional Decomposition. Discussion follows in the end.

\subsection{Data Collection and Pre-processing}

\begin{figure}[H]
    \centering
    \begin{tabular}{ c }
        \includegraphics[width=0.38\linewidth]{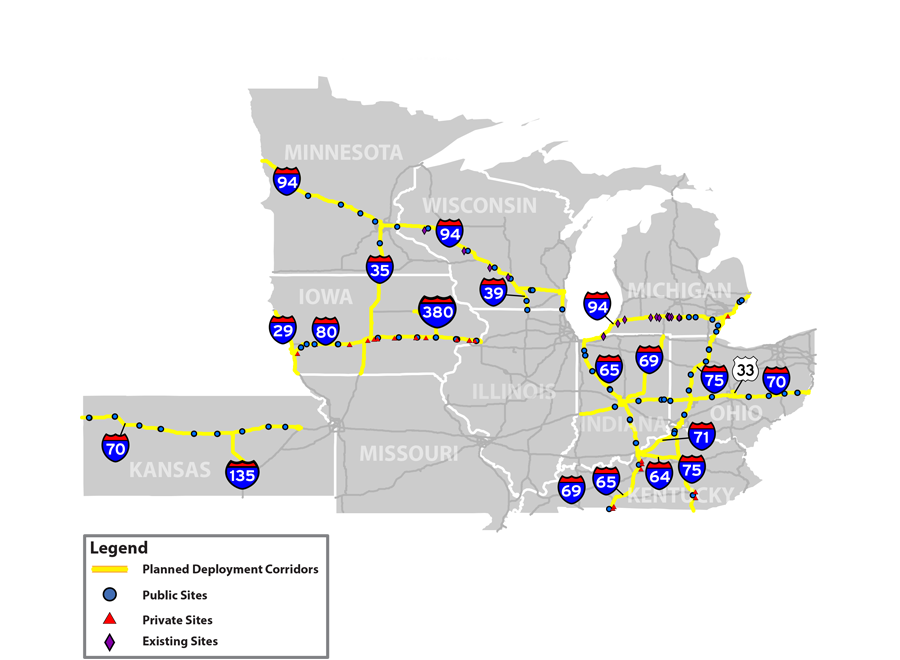}
    \end{tabular}
    \begin{tabular}{ c c }
        \includegraphics[width=0.50\linewidth]{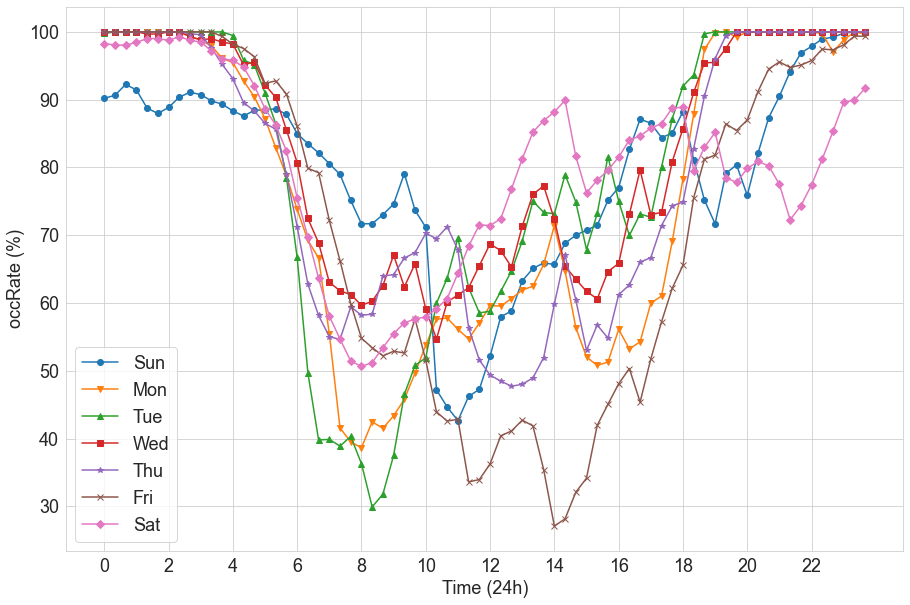} \\
        
        \includegraphics[width=0.50\linewidth]{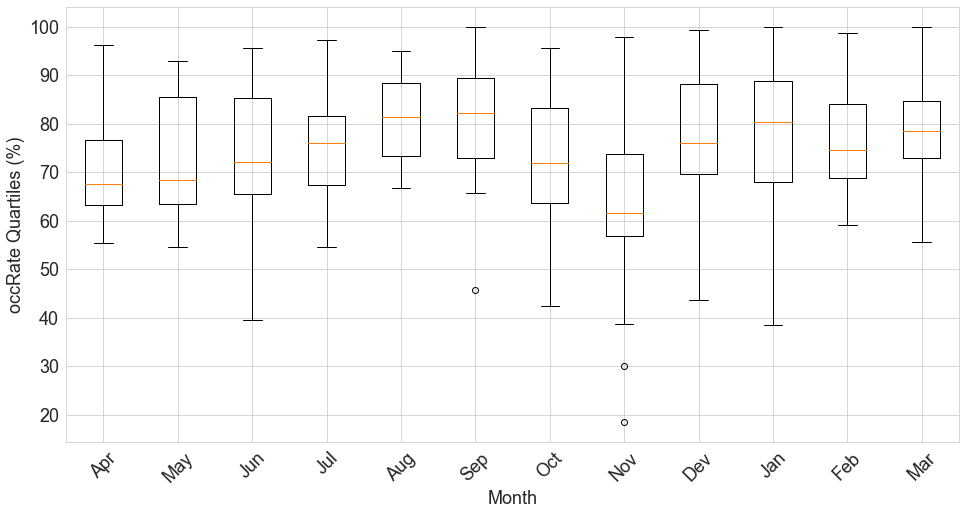}
    \end{tabular}
    \caption{Spatial distribution of truck parking sites in eight states in MAASTO (left). Average hourly occupancy rate at one site in a week averaged by days of 4 weeks (right top). Monthly whiskers data visualization (right bottom) \cite{rei2023}.}\label{fig:truck_parking}
\end{figure}

Figure \ref{fig:truck_parking} (Left) shows all truck parking sites in MAASTO states and where we have collected parking data. Currently (as of Oct, 2023), TPIMS-DAS has archived data from 144 parking sites (public and private) of the eight participating MAASTO states. The number of current sites reporting data in each state is Iowa: 44, Illinois: 19, Kansas: 18, Kentucky: 13, Michigan: 14, Minnesota: 7, Ohio: 18, and Wisconsin: 11. About 160,000 parking records are archived in TPIMS-DAS daily; hence more than 1000 records per site on average. Due to maintenance periods in some sites and a lack of consequent data, we picked 104 sites for the graph construction and usage prediction. Right figures on Figure \ref{fig:truck_parking} show average occupancy rate in daily and monthly from our previous research \cite{rei2023}.

To apply TPIMS data in our model, we pre-processed the raw data in the appropriate format. We extracted dependent features from archived data following the formats in Table \ref{tab:desc_feature}. Week ID, day ID, and hour ID are dynamically changed, and others are static features corresponding to each site. These features were identity encoded and used as the input features of the nodes. The occupancy rate is calculated using the available data and capacity and is the target feature of the model. 

Edges connect two different sites, and nodes are partially connected with the characteristics of highway networks. Following that, we selected sites within 40 miles, which is approximately a 35-minute driving distance, then connected them to construct single connected graph and regional subgraphs. Each edge has the actual driving distance as weights extracted from Bing Maps Locations API. For this research, we used 10-minute frequency data, while some sites have different data collection frequencies. Therefore, we interpolated the occupancy rate data with the linear interpolation. Other dynamic features were determined to be the same number as the previous timestep. Please refer to more details from here (\url{https://github.com/raynbowy23/TPIMSDataset}).

\begin{table}[h]
\caption{Feature descriptions on our dataset.}
\centering
\resizebox{\linewidth}{!}{\begin{tabular}{c|l}
\hline
\rule{0pt}{3ex} \textbf{Feature} & \multicolumn{1}{c}{\textbf{Description}}                                              \\ \hline
\rule{0pt}{3ex} Week ID          & Week ID labeled for each week from 1                              \\ \hline
\rule{0pt}{3ex} Day ID           & Day ID labeled for each day                                       \\ \hline
\rule{0pt}{3ex} Hour ID          & Hour ID labeled for each hour (0-23)                              \\ \hline
\rule{0pt}{3ex} Travel Time      & Travel time from neighbor city to the site                        \\ \hline
\rule{0pt}{3ex} Owner            & 1 if the public owner runs the site, and 0 for private                \\ \hline
\rule{0pt}{3ex} Amenity          & The number of amenities (e.g., restrooms, handicap accessible, Wi-Fi) \\ \hline
\rule{0pt}{3ex} Capacity         & The number of maximum availability of parking lots                \\ \hline
\rule{0pt}{3ex} Occupancy Rate   & The percentage of used parking lots against capacity              \\ \hline
\end{tabular}}
\label{tab:desc_feature}
\end{table}

\subsection{Evaluation Metrics}
Usage prediction metrics compare the predicted value and ground truth data statistics. They are calculated using the root mean squared error (RMSE), mean absolute error (MAE), and mean absolute percentage error (MAPE). We used the occupancy rates on every site for the target value, which is the parking usage divided by the full capacity. The distance between the predicted and ground truth values is interpreted as the prediction accuracy. In addition, RMSE also considers the outliers by providing a large penalty.

\begin{equation}
\begin{split}
    \text{RMSE} &= \sqrt{\frac{1}{N} \sum_{n=1}^N\sum_{t=1}^T (\mathbf{O}_{n_t} - \mathbf{\hat{O}}_{n_t})^2} \\
    \text{MAE} &= \frac{1}{N} \sum_{n=1}^N\sum_{t=1}^T |\mathbf{O}_{n_t} - \mathbf{\hat{O}}_{n_t}|^2 \\
    \text{MAPE} &= \frac{1}{N} \sum_{n=1}^N\sum_{t=1}^T |\frac{(\mathbf{O}_{n_i} - \mathbf{\hat{O}}_{n_t})^2}{q_{95}(\mathbf{O}_n)}|
\end{split}
\end{equation}

$N$ is the number of truck parking sites, and $q_{95}$ is the 95th percentile of the approximate ground truth value as a "reference capacity" \cite{yang2019}. Due to the working time limitation and constraints of information distribution, over-usage of the maximum capacity is sometimes observed in some truck parking sites. Hence, we utilized the 95th percentile occupancy to represent each site's parking capacity. 

\subsection{Model Comparison}

We compared our proposed framework, Regional Temporal GCN (RegT-GCN), with the following baseline methods: Stacked GRU, Stacked GCN \cite{defferrard2016}, Temporal GCN (T-GCN) \cite{bai2021}, Temporal Graph Convolution LSTM (LTGC) \cite{seo2018}, Graph SAGE \cite{hamilton2017}, Spatial and Temporal Normalization (STNorm) \cite{deng2021}, and Spatial Temporla Identity (STID) \cite{shao2022}. They were all applied RMSProp and the learning rate is $10^{-3}$ with the decaying rate of $10^{-4}$. Input channels are 8 for all models and output channels are 512 for LSTM, T-GCN, and LTGC and 256 for the other models. For this experiment, we used one GeForce RTX 3080 for training and inference. Experimentally, our model takes around 15 minutes for training and 1 minute for inference.

\begin{itemize}
    
    \item \textbf{Stacked GRU} is a model deliberating temporal dependencies. Empirically, two layers of GRU perform better than one or three layers in this task setting.

    \item \textbf{Stacked GCN} is a model deliberating spatial dependencies. The model composes two layers of GCN \cite{defferrard2016}.
    
    \item \textbf{T-GCN} is composed of GRU as a temporal module with attention-based aggregation (A3TGCN) on its spatial module \cite{bai2021}.

    \item \textbf{LTGC} is based on Chebyshev Graph Convolutional Long Short Term Memory Cell \cite{seo2018} with 2 for Chebyshev filter size and the symmetric normalization.    
    

    \item \textbf{Graph SAGE} \cite{hamilton2017} is a general framework for inductive representation learning on large graphs with low-dimensional node embeddings.

    \item \textbf{STNorm} \cite{deng2021} serves as a normalization module designed for temporal and spatial features. It employs WaveNet \cite{oord2016} as its backbone model to effectively separate and refine the high-frequency components and local patterns embedded within the raw data.
    
    \item \textbf{STID} \cite{shao2022} integrates spatial and temporal identities to enhance predictive performance.

    \item \textbf{RegT-GCN} uses A3TGCN for the spatial module and GRU for the temporal module. Input subgraphs are created with Regional Decomposition.
    
\end{itemize}

\subsection{Truck Parking Usage Prediction}

\begin{table*}[!t]
\caption{The comparison of prediction results using baseline and our models on the truck parking dataset with different time horizons (displayed in minutes).}
\resizebox{\textwidth}{!}{\begin{tabular}{lcccccccccccc} \hline 

\rule{0pt}{3ex}                                & \multicolumn{4}{c}{RMSE}& \multicolumn{4}{c}{MAE}& \multicolumn{4}{c}{MAPE}\\ 
& 10 & 30 & 120 & 360 & 10 & 30 & 120 & 360 & 10 & 30 & 120 & 360 \\  \hline  

\rule{0pt}{3ex} STNorm \cite{deng2021} & 0.119 & 0.146 & 0.184 & 0.158 & 0.092 & 0.116 & 0.142 & 0.122 & 13.12 & 16.36 & 19.64 & 16.91 \\ 
\rule{0pt}{3ex} Stacked GRU & 0.116 & 0.134 & 0.165 & 0.177 & 0.090 & 0.105 & 0.128 & 0.138 & 12.48 & 14.97 & 17.84 & 17.54 \\
\rule{0pt}{3ex} Stacked GCN \cite{defferrard2016}& 0.114 & 0.120 & 0.131 & 0.159 & 0.088 & 0.093 & 0.099 & 0.123 & 12.23 & 13.18 & 14.10 & 16.74 \\
\rule{0pt}{3ex} LTGC \cite{seo2018} & 0.105 & 0.098 & 0.130 & 0.158 & 0.082 & 0.072  & \textbf{0.082} & 0.122 & 11.25 & 10.05  & 13.65 & 16.60 \\


\rule{0pt}{3ex} Graph SAGE \cite{hamilton2017} & 0.102 & 0.112 & 0.132 & 0.155 & 0.078 & 0.087 & 0.101 & 0.119 & 11.37 & 11.93 & 13.91 & 16.33 \\
\rule{0pt}{3ex} T-GCN \cite{bai2021} & 0.101 & 0.115 & 0.138 & 0.164 & 0.076 &  0.089 & 0.107  & 0.127 & 11.08 & 12.78 & 15.28 & 17.25 \\ 
\rule{0pt}{3ex} STID \cite{shao2022} & \textbf{0.082} & 0.097 & 0.123 & 0.153 & \textbf{0.060} & 0.073 & 0.093 & 0.119 & \textbf{8.31} & 10.07 & 12.88 & 16.47 \\ 
\hline
\rule{0pt}{3ex} \textbf{RegT-GCN} (ours)  & 0.086 & \textbf{0.094} & \textbf{0.120} & \textbf{0.150} & 0.062 & \textbf{0.067} & 0.088 & \textbf{0.114} & 8.76 & \textbf{9.36} & \textbf{12.43} & \textbf{15.68} \\
\hline 
\end{tabular}}
\label{tab:res_table}
\end{table*}
\begin{figure*}[t]
    \centering
    \includegraphics[width=\textwidth]{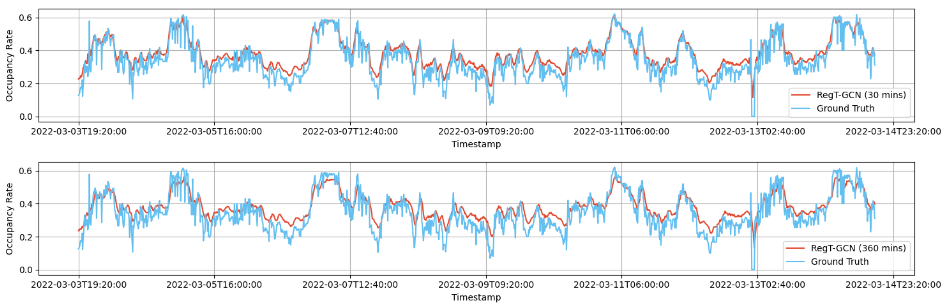}
    \caption{\label{fig:reg_decomp} The results of occupancy rate prediction with RegT-GCN at one truck parking site for the time horizon of 30 minutes (above) and 360 minutes (below) and ground truth from Mar. 3rd, 2022 to Mar. 14th, 2022.}
\end{figure*}


Each model was trained on $20\%$ of the data (March 1st to March 3rd, 2022) and tested on the rest of $80\%$ of the data (March 3rd to March 14th, 2022). Table \ref{tab:res_table} demonstrates the performance comparison with baseline models. Comparing the temporal and spatial models, Stacked GCN predicted generally better than Stacked GRU, which indicates the spatial model learns better representations of truck parking sites than the temporal model. In comparison, most spatio-temporal models improved the model performance, especially in shorter time horizons compared to Stacked GRU and Stacked GCN. Across the models, our RegT-GCN outperforms the baselines. Specifically in longitudinal analysis, Stacked GRU decreased its performance compared to Stacked GCN since the temporal model highly depends on the predicted time horizons. At longer time horizons, the result of Stacked GCN predicted a better or closer accuracy to spatio-temporal models, while Stacked GRU, T-GCN, and STID performance dropped the prediction accuracy. Across most time horizons, our RegT-GCN successfully demonstrates improvements in truck parking prediction performance. 

Figure \ref{fig:reg_decomp} illustrates the predicted results of RegT-GCN and ground truth. The above figure compares the predicted occupancy rate over the timestamp beginning from Mar. 3rd, 2022. The prediction results show that the model successfully learns the tendencies of truck parking usage, which also adequately predicts intense usage. The figure below shows the predicted occupancy rate over a time horizon of 360 minutes. RegT-GCN also learns the tendency of the usage at whole timestamps. However, compared to a time horizon of 30 minutes, RegT-GCN exhibits limitations in predictive performance and capturing peak values accurately. These examples prove that graph decomposition can be effective in any time horizon, such that truck drivers can schedule their route anytime.
\subsection{Model Generality}

\begin{table*}[!t]
\caption{The generalizability comparison of prediction results using baseline and our models on the truck parking dataset with different time horizons (displayed in minutes).}
\resizebox{\textwidth}{!}{\begin{tabular}{lcccccccccccc} \hline 

\rule{0pt}{3ex}                                & \multicolumn{4}{c}{RMSE}& \multicolumn{4}{c}{MAE}& \multicolumn{4}{c}{MAPE}\\ 
 & 10 & 30 & 120 & 360 & 10 & 30 & 120 & 360 & 10 & 30 & 120 & 360 \\  \hline  

\rule{0pt}{3ex} STNorm \cite{deng2021} & 0.126 & 0.164 & 0.206 & 0.160 & 0.098 & 0.130 & 0.160 & 0.124 & 12.95 & 17.01 & 20.58 & 15.74 \\ 
\rule{0pt}{3ex} Stacked GRU & 0.119 & 0.129 & 0.167 & 0.177 & 0.095 & 0.100 & 0.130 & 0.138 & 12.33 & 13.27 & 16.78 & 17.54 \\
\rule{0pt}{3ex} Stacked GCN \cite{defferrard2016} & 0.114 & 0.127 & 0.129 & 0.164 & 0.088 & 0.096 & 0.098 & 0.128 & 12.23 & 12.71 & 12.83 & 16.34 \\
\rule{0pt}{3ex} LTGC \cite{seo2018} & 0.114 & 0.102 & 0.139 & 0.172 & 0.092 & 0.078  & 0.110 & 0.136 & 11.95 & 10.13  & 14.04 & 17.24 \\
\rule{0pt}{3ex} Graph SAGE \cite{hamilton2017} & 0.100 & 0.118 & 0.138 & 0.159 & 0.076 & 0.095 & 0.108 & 0.124 & 10.31 & 12.20 & 13.88 & 15.72 \\
\rule{0pt}{3ex} T-GCN \cite{bai2021} & 0.097& 0.109 & 0.129 & 0.171 & 0.072 & 0.083 & 0.099 & 0.135 & 9.76 & 11.12 & 13.05 & 17.15 \\ 
\rule{0pt}{3ex} STID \cite{shao2022} & \textbf{0.083} & 0.108 & 0.128 & 0.149 & 0.062 & 0.082 & 0.099 & 0.116 & 8.14 & 10.65 & 12.74 & 14.79 \\ 
\hline 
\rule{0pt}{3ex} \textbf{RegT-GCN} (ours)  & 0.085 & \textbf{0.092} & \textbf{0.117} & \textbf{0.148} & \textbf{0.061} & \textbf{0.066} & \textbf{0.087} & \textbf{0.113} & \textbf{8.12} & \textbf{8.73} & \textbf{11.30} & \textbf{14.46} \\
\hline 
\end{tabular}}
\label{tab:res_gen_table}
\end{table*}

To assess the generalizability of the models, we selected additional weeks, starting from March 15th, 2022, for the test sequences and conducted inference experiments with the pretrained models. Table \ref{tab:res_gen_table} summarizes the results of RMSE, MAE, and MAPE to evaluate the model's predictive accuracy. In the table, every model shows the general capability of its predictive abilities to other time ranges.

Further analysis reveals that RegT-GCN consistently outperformed baseline models across various time horizons, indicating their superior predictive capabilities. Remarkably, even when baseline models exhibit lower performance than the original prediction results, models with integrated decomposition techniques maintain their efficacy in providing accurate and generalizable inferences. These findings underscore the consistent and superior performance of the Regional Decomposition in predictive accuracy, regardless of the time horizon under consideration.






\begin{figure}[!ht]
    \centering
    \includegraphics[width=\linewidth]{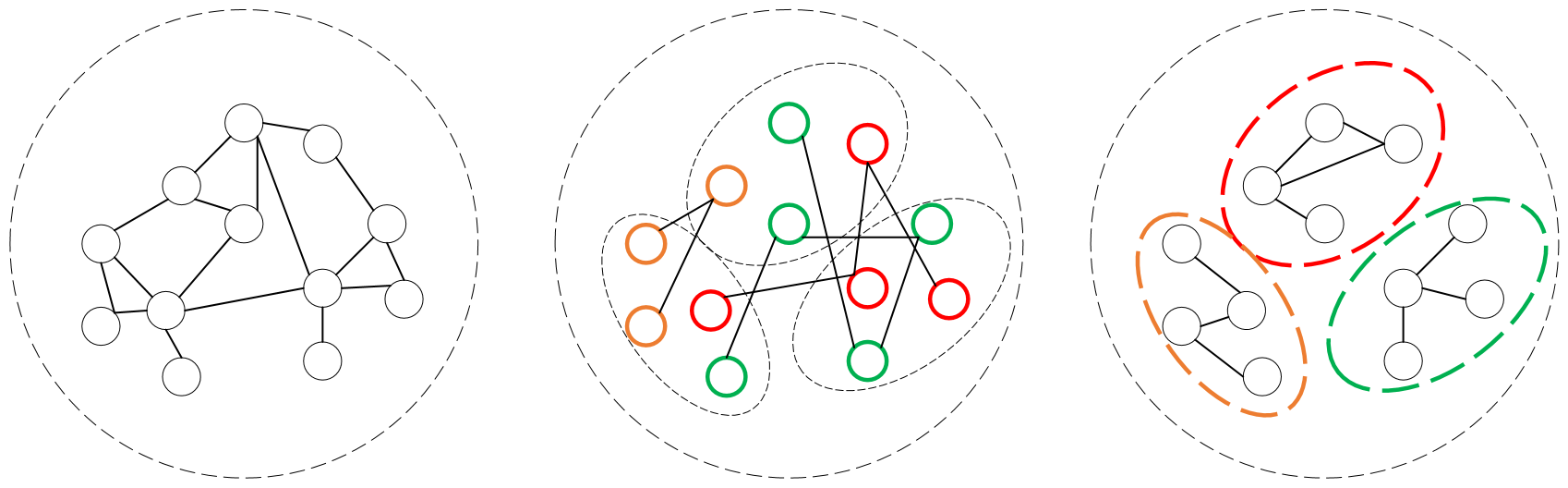}
    \caption{Variety of connectivity on the simple graphs (Left: connected, Middle: randomly connected, Right: regionally connected). Small ellipses represent regional classification. The same color denotes the same group constructing one graph.}\label{fig:relation}
\end{figure}

\subsection{Regional vs Random Graph Decomposition}

We also conducted an experiment on graph connectivity exploration. Figure \ref{fig:relation} illustrates the diversity of connection approaches. We divided the truck parking location graph into subgraphs characterized by random connections. The proposed model, Random T-GCN (RanT-GCN), is analogous to DropEdge \cite{rong2019}, which applied random edge sampling on the graph to make several subgraphs. Most modules are the same as RegT-GCN, but use random connected graphs as input subgraphs. RanT-GCN separates an original graph into a subgraph, which possesses the same number of nodes in each region with the graph decomposition, and each node is randomly selected and connected together. We also implemented the variants of RanT-GCN.\begin{itemize}
    \item P: The total number of nodes in each region was randomly selected.
    \item R: Each node was randomly selected to connect to each other.
    \item S: Set the maximum degree of selected nodes to four.
\end{itemize}

\begin{table}[h]
\caption{Different Connected Graph-Based Model Comparison at Time Horizon 10}
\small
\centering
\begin{tabular}{lccc} \hline 
 \rule{0pt}{3ex} & RMSE & MSE & MAPE \\ \hline
 \rule{0pt}{3ex} RanT-GCN & 0.090 & 0.066 & 9.124  \\
 \rule{0pt}{3ex} RanT-GCN (R) & 0.090 & 0.066 & 9.125  \\
 \rule{0pt}{3ex} RanT-GCN (P) & 0.089 & 0.065 & 9.080 \\
 \rule{0pt}{3ex} RanT-GCN (S) & 0.089 & 0.064 & 9.14 \\
 \hline
 \rule{0pt}{3ex} \textbf{RegT-GCN} & 0.086 & 0.062 & 8.75 \\

\hline
\end{tabular}
\label{tab:connectivity}
\end{table}


In the study of different connectivity approaches, RanT-GCN and RegT-GCN outperform the baseline models in most of the prediction time horizons. Within these two, regionally decomposing a graph achieved the best accuracy. Table \ref{tab:connectivity} explains that node selection contributes more than changing the degree of nodes and regional selection. However, the experiments on different node selections (RanT-GCN vs. RanT-GCN (R)) came to the same results; thus, the node selection should be taken carefully, as we will mention in our discussion.

\begin{figure}
    \centering
    \begin{subfigure}{0.8\linewidth}
        \includegraphics[width=\linewidth]{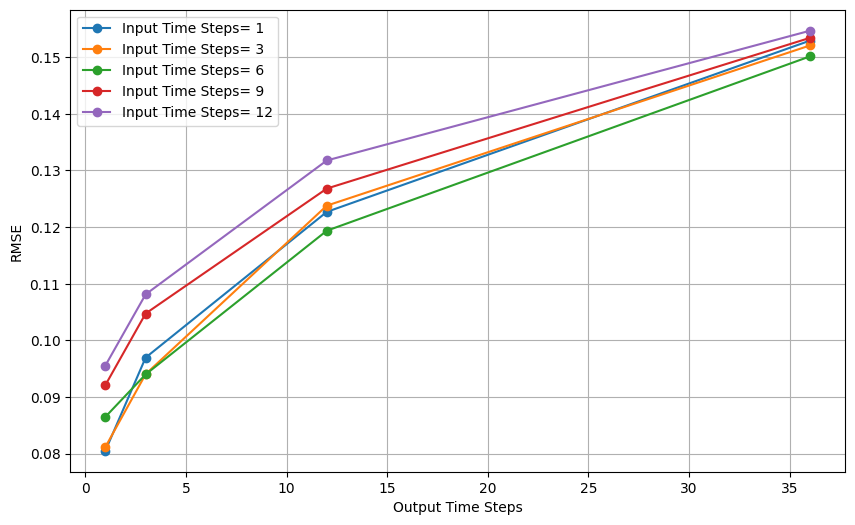}
    \end{subfigure}
    \begin{subfigure}{0.45\linewidth}
        \includegraphics[width=\linewidth]{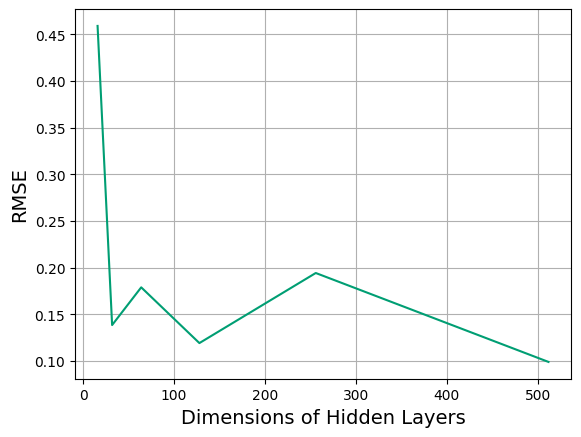}
    \end{subfigure}
    \begin{subfigure}{0.45\linewidth}
        \includegraphics[width=\linewidth]{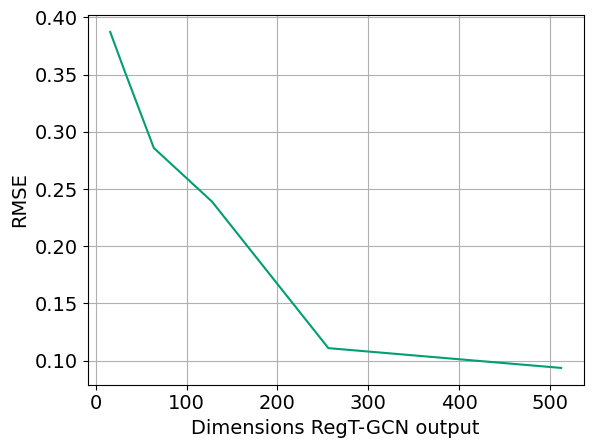}
    \end{subfigure}
    \caption{Sensitivity analysis of varying input time steps (top), hidden layers dimensions (bottom left), and output dimensions of RegT-GCN (bottom right).}
    \label{fig:sensitivity}
\end{figure}

\subsection{Parameters Sensitivity Analysis}
Lastly, we examined the model sensitivity by varying the input time step as \{1, 3, 6, 9, 12\}, the hidden layers dimension as 
\{16, 32, 64, 128, 256, 512\}, and the output dimension of RegT-GCN as \{16, 32, 64, 128, 256, 512\} (Figure \ref{fig:sensitivity}). When testing the hidden layer dimensions, the output dimensions were fixed at 256, and vice versa.

The first analysis (top) demonstrates that shorter input time steps initially produce lower RMSE, and the results are not always consistent across different output horizons, although longer input sequences are often expected to yield better predictive performance. Interestingly, the RMSE values for all input configurations converge as the output time steps increase. This convergence indicates that, as the model predicts farther into the future, the accumulated prediction uncertainty limits the impact of the initial input length. Similarly, we can assume that increasing the input length yields diminishing results at a certain point when the input length is approximately balanced.

The second analysis (bottom left) indicates that while increasing hidden layer dimensions generally reduces RMSE, there are fluctuations at 256 dimensions. Finally, the third analysis (bottom right) shows that increasing the RegT-GCN output dimensions improves the accuracy, with RMSE decreasing up to 256 dimensions, after which the improvements plateau. While larger output spaces allow richer feature representations, there is a point beyond which additional complexity yields diminishing returns. These results suggest that model architecture must be carefully tuned to balance performance, as overly complex networks may not always improve results.

\subsection{Discussion}

\subsubsection{Computational Efficiency}
The superior performance of our models stems from a reduced number of overlaps among reference nodes. Specifically, in a connected graph with self-loops, each node is calculated $N * l$ times, where $N$ expresses the number of nodes and $l$ represents the number of neighbors being sampled for each node. However, these overlaps hinder the learning of the graph structure since a node requires more computation to refer to more nodes. In contrast, our model addresses this issue by creating subgraphs and reducing the overlap to $N^r * l^r$ for each spatial module, where $N^r$ and $l^r$ denote the number of nodes and neighboring nodes in subgraphs. As a result, the model can prioritize the crucial truck parking feature in the region through simple calculations.

\subsubsection{Graph Connectivity}
The experiment results show that both the regional and random decomposition techniques outperform the non-decomposed method, T-GCN. This suggests that the concatenation of hidden vectors plays a crucial role in sharing subgraph features, highlighting the importance of employing graph neural networks in conjunction with pre-clustering graphs for superior performance. 

At the same time, the connectivity of the decomposed graph needs to be taken into account. According to the analysis of previous research \cite{rei2023}, truck drivers tend to use consecutive parking sites when one of them is fully occupied. Consequently, considering cooperative node interactions within regions yields better outcomes. Hence, the competitive performance of the Regional Decomposition approach in comparison to a single or random connected graph indicates that our framework can achieve strong performance simply by decomposing the input graph.

\subsection{Limitations and Future Work}
\noindent\textbf{Route Planning and Graph Connectivity:} Corresponding to the truck driver's decision-making process, our research is limited to exploring the graph connection along with their route planning. We will explore the concept of decomposition by their route planning network in our future research.

\noindent\textbf{Over-smoothing:} The investigation of the limits of over-smoothing \cite{oono2020} on our models is also important. By incorporating deeper layers and exploring the potential for extending the prediction time horizon, the proposed model's maximum allowable performance limits should be investigated in the future.

\noindent\textbf{Model Composability:} It is necessary to design an end-to-end framework capable of automatically decomposing graphs and integrating spatio-temporal modules for inductive scenarios. This will enable us to apply the models to previously unseen graphs, such as those outside the scope of participating states in MAASTO, while maintaining consistent and robust performance.

\section{Conclusion}

In this paper, we propose the Regional Temporal Graph Convolutional Network (RegT-GCN), a state-wide truck parking usage prediction framework, leveraging topological structures of truck parking site distributions. The RegT-GCN comprises a Regional Decomposition approach to learn the geographical characteristics efficiently and a spatio-temporal module to capture the temporal dependencies on graphs. Experiments using our original field dataset show that the proposed models outperform other baseline models. Decomposed graphs into regional-oriented subgraphs result in better prediction capability and consistency of inferences. The study also encompasses a demonstration of predicting usages through the empirical observation of non-informative loss, facilitated by the integration of the Gated Recurrent Unit module, which adeptly adapts to value fluctuations. Furthermore, with extensive experiments focused on graph connectivity, this paper validates that incorporating regional relationships of truck parking enhances the comprehension of the structural dynamics of truck parking behavior.

\bibliography{reference}{}
\bibliographystyle{IEEEtran}

\section{Biography Section}

\begin{IEEEbiography}[{\includegraphics[width=1in,height=1.25in,clip,keepaspectratio]{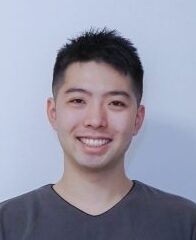}}]{Rei Tamaru} received his Bachelor of Arts degree in Information Science from International Christian University, Tokyo, Japan, in 2019. He received a Master of Philosophy in Transdisciplinary Sciences from Japan Advanced Institute of Science and Technology in 2021. He is currently a Ph.D. student in Civil and Environmental Engineering at the University of Wisconsin-Madison. His research directions are intelligent transportation systems, connected automated vehicles, and traffic simulations.
\end{IEEEbiography}

\begin{IEEEbiography}[{\includegraphics[width=1in,height=1.25in,clip,keepaspectratio]{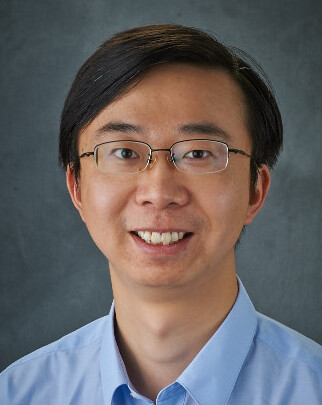}}]{Yang Cheng} received the B.S. and M.S. degrees in automation from Tsinghua University, Beijing, China, in 2004 and 2006, respectively, and the Ph.D. degree in civil engineering from the University of Wisconsin–Madison in 2011. He is currently a scientist at the Wisconsin Traffic Operations and Safety (TOPS) Laboratory of the University of Wisconsin-Madison (UW). His research areas include automated highway and driving systems, mobile traffic sensor modeling, large-scale transportation data management and analytics, and traffic operations and control.
.\end{IEEEbiography}

\begin{IEEEbiography}[{\includegraphics[width=1in,height=1.25in,clip,keepaspectratio]{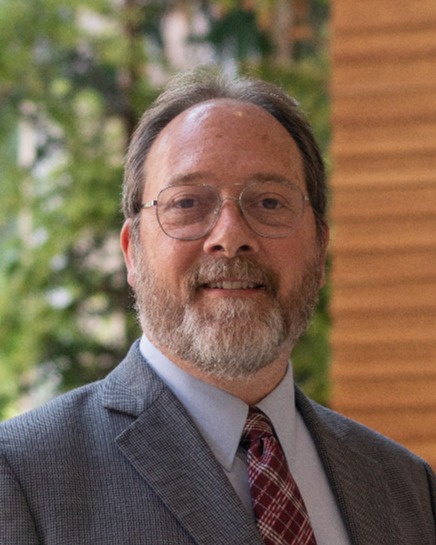}}]{Steven T. Parker} is the Managing Director of the Wisconsin Traffic Operations and Safety (TOPS) Laboratory at the University of Wisconsin-Madison. In this capacity, he has led a range of research and development initiatives for the TOPS Lab across several core areas including transportation safety, work zone systems, traffic management systems, and connected and automated vehicle technologies. Dr. Parker has over 25 years of professional experience in applied research computing with the last two decades working on transportation systems and technology issues in collaboration with the Wisconsin Department of Transportation (WisDOT) and other agency partners. Prior to joining the TOPS Lab, Dr. Parker received a Ph.D. in Computer Science from the University of Wisconsin-Madison. He is currently serving in his second term as the Chair of the Transportation Research Board (TRB) AED30 Information Systems and Technology Committee
\end{IEEEbiography}

\begin{IEEEbiography}[{\includegraphics[width=1in,height=1.25in,clip,keepaspectratio]{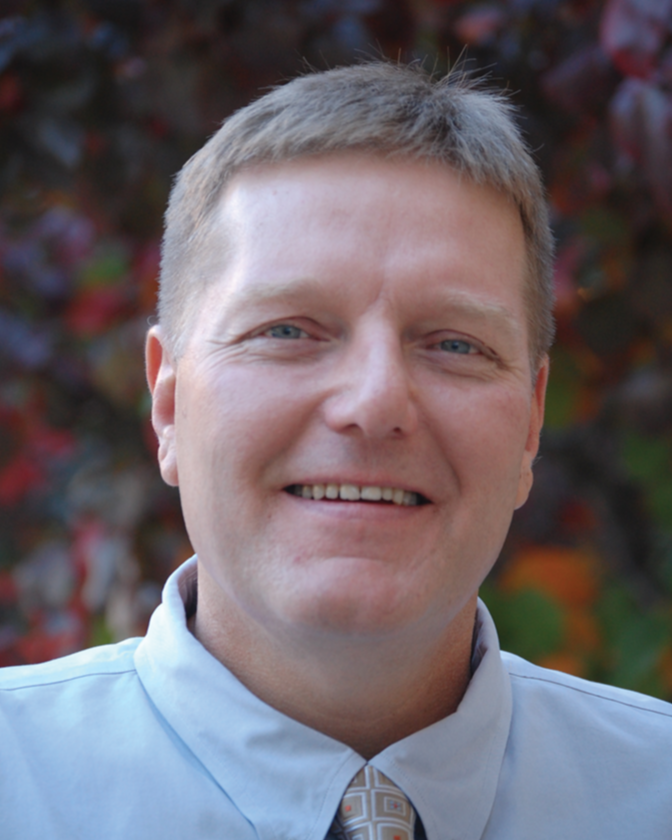}}]{Ernie Perry} is the senior researcher and facilitator at the Mid America Freight Coalition (MAFC).  He directs the multimodal freight policy, programming, and operation research and collaboration for the 10 states of the Mid America Association of State Transportation Officials. He is an expert in multimodal freight systems and planning, multimodal freight operations, economic development, and multistate collaboration. Perry has completed over 30 projects with the Coalition in a wide range of areas including truck parking, truck electrification, marine planning, aviation planning and economic impacts, freight data, over size and over dimension loads, multistate freight planning, and the value of multimodal freight movements.
\end{IEEEbiography}

\begin{IEEEbiography}[{\includegraphics[width=1in,height=1.25in,clip,keepaspectratio]{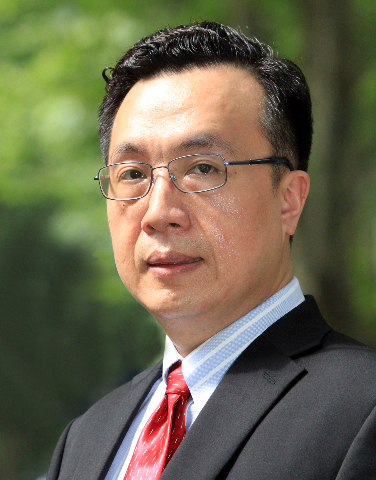}}]{Bin Ran} is the Vilas Distinguished Achievement Professor and Director of ITS Program at the University of Wisconsin, Madison. He is an expert in dynamic transportation network models, traffic simulation and control, traffic information system, Internet of Mobility, Connected Automated Vehicle Highway (CAVH) System. He has led the development and deployment of various traffic information systems and the demonstration of CAVH systems. He is the author of two leading textbooks on dynamic traffic networks. He has co-authored more than 240 journal papers and more than 260 referenced papers at national and international conferences. He holds more than 20 patents of CAVH in the US and other countries. He is an associate editor of Journal of Intelligent Transportation Systems.
\end{IEEEbiography}

\begin{IEEEbiography}[{\includegraphics[width=1in,height=1.25in,clip,keepaspectratio]{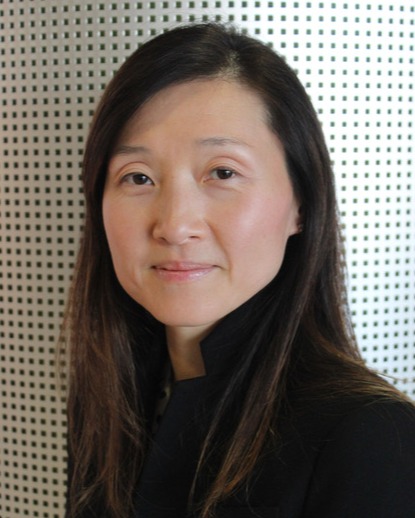}}]{Soyoung (Sue) Ahn}
is a professor in civil and environmental engineering at the University of Wisconsin-Madison and an executive director of the Mid America Freight Coalition (MAFC). Her recent research involves (i) analysis and modeling of traffic flow mixed with connected autonomous vehicles (CAVs), (ii) development of CAV platoon control strategies, (iii) development of system control strategies using CAVs, and (iv) freight transportation planning. She is a Senior Editor for IEEE Transactions on ITS, and an Associate Editor for Transportation Research Part C and Transportation Research Record. She also serves as an editorial board editor for Transportation Research Part B. She is a chair of the Operations Section of Transportation Research Board and an elected member of the International Advisory Committee for the International Symposium on Traffic and Transportation Theory.\end{IEEEbiography}

\vfill

\end{document}